\documentclass{esannV2}
\usepackage{graphicx}
\usepackage[latin1]{inputenc}
\usepackage{amssymb,amsmath,array}
\usepackage{multirow}
\usepackage{tabularx, booktabs, makecell, caption}
\usepackage{siunitx}
\usepackage{hyperref}
\usepackage{xurl}

\begin{document}
\title{Lifelong Learning from Event-based Data}

\author{Vadym Gryshchuk$^1$, Cornelius Weber$^1$, Chu Kiong Loo$^2$, and Stefan Wermter$^1$
%
\thanks{This work was supported in part by the Georg Forster Research Fellowship for Experienced Researchers (Neurosynthelligence) from the Alexander von Humboldt Foundation, and in part by the German Research Foundation (DFG) on Crossmodal Learning under Grant TRR169.}
%
\vspace{.3cm}\\
%
1- University of Hamburg - Department of Informatics \\
Vogt-Koelln-Strasse 30, 22527 Hamburg - Germany
%
\vspace{.1cm}\\
2- University of Malaya - Department of Artificial Intelligence \\Lembah Pantai, 50603 Kuala Lumpur - Malaysia\\
}

\maketitle

\begin{abstract}
Lifelong learning is a long-standing aim for artificial agents that act in dynamic environments, in which an agent needs to accumulate knowledge incrementally without forgetting previously learned representations. We investigate methods for learning from data produced by event cameras and compare techniques to mitigate forgetting while learning incrementally. We propose a model that is composed of both, feature extraction and continuous learning. Furthermore, we introduce a habituation-based method to mitigate forgetting. Our experimental results show that the combination of different techniques can help to avoid catastrophic forgetting while learning incrementally from the features provided by the extraction module.
\end{abstract}

\section{Introduction}
An event camera is a dynamic vision sensor that responds to the changes of brightness at any pixel location. Lower power consumption, higher dynamic range and high temporal resolution are the advantages of event cameras over conventional shutter cameras. These advantages make an event sensor suitable for real-life applications that rely on fast responses, time and scene illumination. Such scenarios build the ground for the development of artificial systems that are dependent on the process of continuous learning. These aspects motivated us to use event-based data for lifelong learning. The architectures for lifelong learning that are based on a pre-trained feature extractor deliver state-of-the-art results \cite{Parisi2018, Ven2020}. However, these architectures consider data that are produced by a conventional shutter camera. In our paper, we show that lifelong learning from event-based data can follow the same strategy. Furthermore, the application of event cameras opens new possibilities for the development of novel learning systems in dynamic environments.

\section{Background}
Since each pixel in an event camera responds to brightness change independently, the generated asynchronous output carries challenges for the processing of such data. The event-based sequences can be processed by event-by-event methods or methods that group events \cite{Gallego2020}. The event-by-event methods process events sequentially. As an example of such a method, Phased LSTM \cite{Neil2016} is an extension to LSTM \cite{Hochreiter1997} and introduces a new time gate. This gate allows the updates to a  memory cell only during some specific periods. Another approach is to group events to image-like data. A histogram is one of the possibilities to convert events to the event frame \cite{Maqueda2018}. To convert events to a histogram, the occurrences of brightness change at any pixel location over a particular period of time are counted. Specifically, the events that refer to a brightness increase are stored in one histogram, and the events that capture brightness decrease are saved to another histogram. Two histograms act then as two input channels to a convolutional neural network (CNN). Since an event camera reacts only to brightness change, a lot of pixel locations in a histogram can contain no values. Thus, a histogram represents the edges of a scene captured by an event camera. However, a conventional convolutional operation causes dilation when the input is sparse. Therefore, a sparse CNN that preserves sparseness is a more reasonable choice \cite{Graham2018}.

Methods that are successfully used to mitigate catastrophic forgetting rely on regularization-based techniques or use replay mechanisms \cite{Parisi2018, Ven2020}. Regularization-based methods restrict the updates to the model's parameters that are important for encoding previous knowledge. One of these methods that estimates this importance is synaptic intelligence \cite{Zenke2017} which introduces a regularization term that is added to the total loss to penalize changes to important parameters while learning a new task. We will present and evaluate a simpler method using a neuron habituation mechanism. The methods that utilize replay mechanisms store either some previous samples or learn the representations for previously learned data. A generative model, in particular, a variational autoencoder can be used to learn latent representations of data \cite{Ven2020}. We show that the architecture for lifelong learning from event-based data can utilize the same methods that are applied to frame-based images.

\section{Approach}
\begin{figure}[!ht]
	\centering
	\includegraphics[scale=0.33]{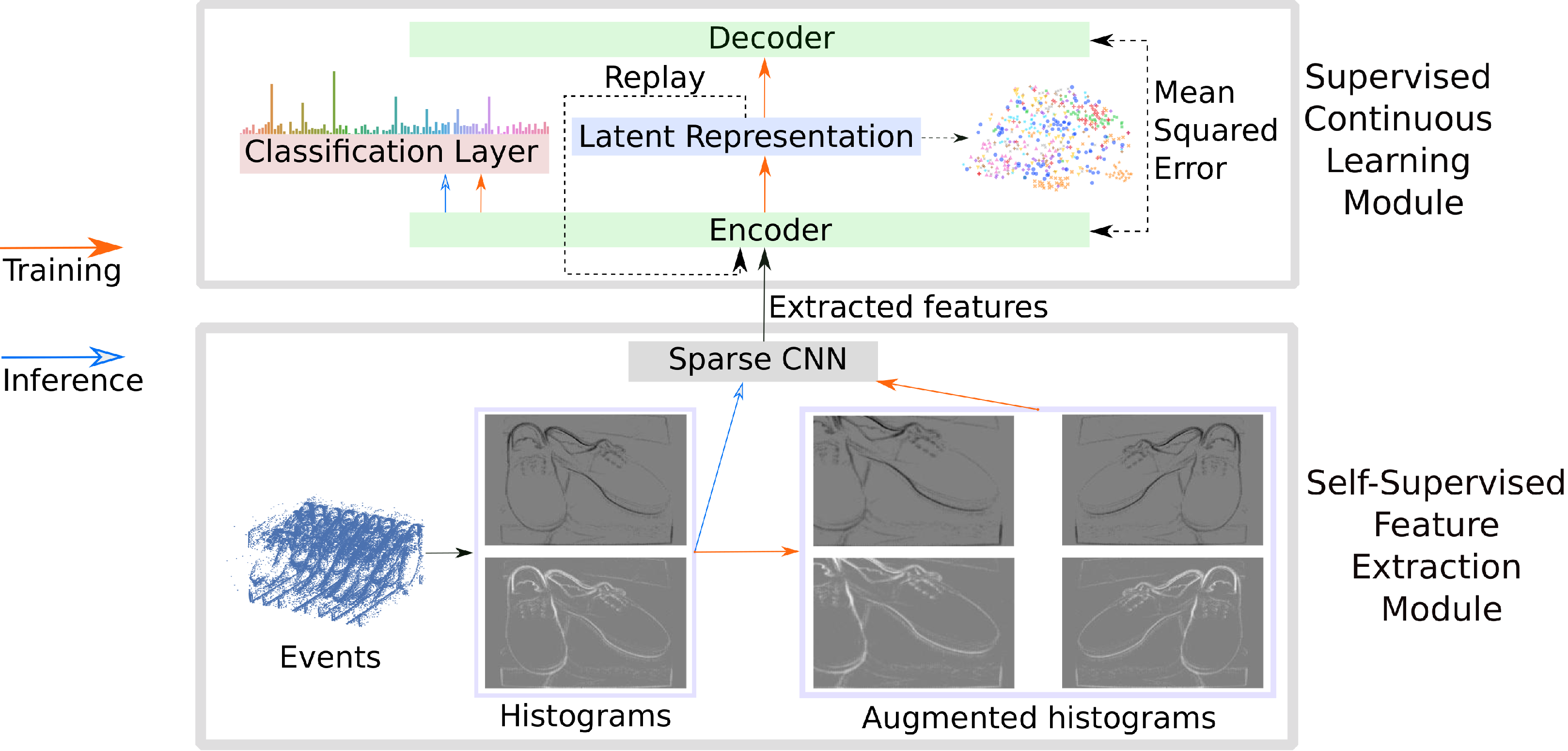}
	\caption{Illustration of the proposed architecture. Sparse CNN trained in a self-supervised way extracts features from the events represented as histograms. The continuous learning component learns incrementally from the extracted features by utilizing replay through the VAE, synaptic intelligence, and habituation.}\label{fig:architecture}
\end{figure}

We propose an architecture that consists of a feature extractor and a component for continuous learning, visualized in Figure~\ref{fig:architecture}. To the best of our knowledge, there are no approaches for direct comparison. Lungu et al. used a memory-based method for incremental learning of hand gestures \cite{Lungu2019}. However, it is questionable if their approach is extensible to scenarios in which the input is more complex.

\subsection{Feature Extraction}

To design a feature extraction module, we compare Phased LSTM and sparse CNN as possible models to extract features from events. We train Phased LSTM in a supervised way and sparse CNN in a self-supervised way following the batch learning strategy. On the one hand, we use Phased LSTM as the event-by-event method, and on the other hand, we utilize Sparse CNN that learns from histograms as grouped events. Based on the results provided in Section~\ref{results}, we select only the self-supervised approach using sparse CNN for the feature extraction module. Self-supervised learning is a subset of unsupervised learning, where no labels are provided during training. We follow the same strategy for self-supervised learning proposed by Chen et al. \cite{Chen2020}. However, instead of frame-based input, we provide events as histograms. The model applies random augmentations directly to histograms, thus learning in a contrastive way by maximizing the agreement between two augmented representations of the same object (Figure~\ref{fig:architecture}, bottom).

\subsection{Continuous Learning}
The module for continuous learning operates on the features provided by the feature extraction module. The learning process follows the incremental strategy, where a model has access only to some object categories during the learning episode. Thus, a learning episode contains only a subset of non-repeating objects that belong to the same category. We base our model on the method proposed by Ven et al. called brain-inspired replay \cite{Ven2020}. It uses a variational autoencoder (VAE) that is trained together with a classifier (Figure~\ref{fig:architecture}, top). 

Additionally, we introduce a habituation-based method to mitigate catastrophic forgetting while learning incrementally. This method utilizes the concept of habituation that was successfully applied to self-organizing neural networks \cite{Parisi2018}. Habituation is the reduction of responses to repeated stimuli. We quip each neuron in the last fully connected layer of an encoder with a habituation counter, which is initialized with $1$. During training, only a part of neurons with the highest activation values are habituated. We slightly modify the habituation update rule presented in \cite{Parisi2018} and define it as follows:

\begin{equation}
\begin{aligned}
\Delta h_i &=  \tau \cdot (1 - h_i) - \tau,
\end{aligned}
\label{eq:habituation_counter}
\end{equation}

where $h_i$ is a habituation counter of a neuron $i$,  $\tau$ is the decay rate that controls a steepness of decay. Consequently, a habituation counter is a monotonically decreasing function. To constrain the changes to the model's parameters, the gradient of a neuron $r_i$ is scaled by the habituation counter of this neuron during each learning iteration:

\begin{equation}
\begin{aligned}
\frac{\partial \mathcal{L}}{\partial r_i} &\gets  \frac{\partial \mathcal{L}}{\partial r_i} \cdot h_i,
\end{aligned}
\label{eq:gradient_h}
\end{equation}

where $\partial$ is a partial derivative and $\mathcal{L}$ is a loss function. Although the habituation-based regularization method is similar to synaptic intelligence, it can be utilized in each layer of a neural network with different values for $\tau$, thus providing more plasticity during learning. We compare different combinations of brain-inspired-replay, synaptic intelligence and habituation to investigate the effect of the habituation-based approach on the mitigation of catastrophic forgetting while learning incrementally.

\section{Experimental Results}\label{results}
We train and evaluate the proposed model on the N-Caltech101 dataset \cite{Orchard2015}. This dataset contains event-based representations from static images. While an image was shown on a screen, an event camera made three saccadic movements to record events. We use the same training and test sets as used in \cite{Messikommer2020}. The code is available from \href{http://software.knowledge-technology.info}{http://software.knowledge-technology.info}. To first evaluate the feature extraction module under optimal conditions, the whole training set is utilized to learn a feature extraction module following the batch learning strategy. Each sample in the dataset can contain dozens of thousands of events, thus the training time of Phased LSTM becomes intractable since Phased LSTM processes events sequentially. Therefore, we randomly select only $5\%$ of events, but at least $5000$. Histograms are created from $50000$ consecutive events; however, this interval of events is randomly placed over the whole sequence of events. Table~\ref{tab:extractor_results} shows the classification accuracy of Phased LSTM and the classification accuracy of a linear classifier trained on top of sparse CNN. Although Phased LSTM achieves worse results, it operates on a portion of events, which can lead to a drop in performance. Yet, using even $5\%$ of events causes a huge overhead in training time when compared to sparse CNN. Furthermore, a feature extractor that is trained in a supervised way on the same training set that is used for the continuous learning module is not a fair condition. Thus, either a feature extractor that is trained without labels or a feature extractor that is used to extract features from different data is a more reasonable approach. Based on these conditions, we use sparse CNN as a feature extractor for the continuous learning module.

\begin{table}[tbp] \centering
	\newcolumntype{C}{>{\centering\arraybackslash}X}
	\sisetup{table-format=1.2, table-number-alignment=center}
	\caption{Evaluation of the feature extraction module on N-Caltech101. \textbf{(left)} Classification accuracy  using Phased LSTM.  \textbf{(right)} Classification accuracy using linear classifier trained on top of frozen features from sparse CNN.}
	\begin{tabular}{lS[table-format=2.2]*{6}{S}}
		\multicolumn{2}{c}{\makecell{Phased LSTM (supervised)}} & \multicolumn{3}{c}{\makecell{sparse CNN (self-supervised)}} \\
		\cmidrule(lr){1-2}\cmidrule(lr){3-5}
		{Training}&{Test}&{Training}&\shortstack{Test Top-1}&\shortstack{Test Top-5}
		\tabularnewline
		\cmidrule[\lightrulewidth](lr){1-2}\cmidrule[\lightrulewidth](lr){3-5}
		35.35&30.90&51.49&42.38&62.60
		\label{tab:extractor_results}
	\end{tabular}
\end{table}

\begin{figure}[h!]
	\centering
	\includegraphics[scale=0.35]{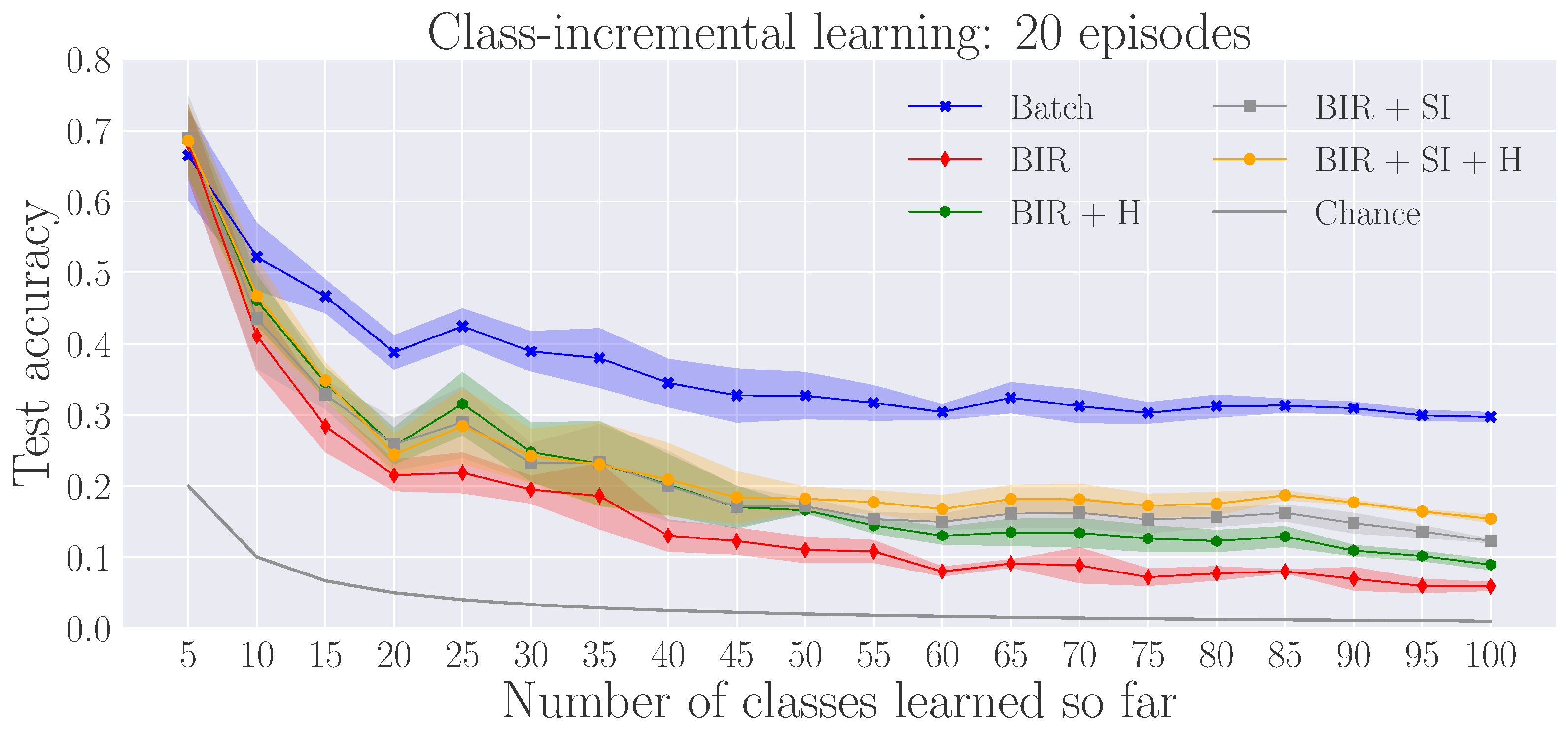}
	\caption{Class-incremental learning on N-Caltech101.}\label{fig:ll_results}
\end{figure}

To evaluate the proposed habituation-based method, we combine habituation with the brain-inspired replay (BIR) and synaptic intelligence (SI) methods. All hyper-parameters are the same for all methods to provide a fair comparison. The strength parameter of SI was found by a grid search and is set to $10^9$. The habituation-based method (H) has two hyperparameters: a decay rate $\tau$ and the fraction $\gamma$ of neurons with the highest activation values that are allowed to be habituated during each learning iteration. We set $\gamma$ to $0.05$ and $0.01$ for the BIR+H and BIR+SI+H methods, respectively. For the strategies BIR+H and BIR+SI+H the values for $\tau$ are set to $0.3$ and $0.02$, respectively. Figure~\ref{fig:ll_results} illustrates class-incremental learning on N-Caltech101. The number of learning episodes is set to 20. Each learning episode contains samples from 5 different non-repeating object categories. The shaded areas show the standard error of the mean. The experiment was executed for three trials, and each trial, a new seed and the random order of classes were used. The BIR+H and BIR+SI methods achieve after learning data of all episodes on average the classification accuracy of $8.94$ and $12.26$, respectively. The addition of the habituation-based method to BIR+SI provides a slight but significant increase in test accuracy: $15.40\pm0.55$.

\section{Conclusion} 
We presented an architecture for lifelong learning, consisting of a feature extractor and a module for continuous learning. We showed that the Phased LSTM is not a favourable method for learning long event-based sequences. The Sparse CNN trained in a self-supervised way achieves better results but histograms discard short time-scale information. A combination of brain-inspired replay and synaptic intelligence with a simple habituation method, which was previously applied to self-organizing neural networks, yields the best performance over class-incremental learning of 100 classes. Furthermore, with this presented approach in this paper we provide useful insights into the application of event cameras for real-life scenarios, in which incremental accumulation of knowledge is crucial.

\begin{footnotesize}

\bibliographystyle{abbrv}
\bibliography{esann}

\end{footnotesize}


\end{document}